\documentclass[10pt,twocolumn,letterpaper]{article}

\usepackage[pagenumbers]{cvpr} 

\usepackage{graphicx}
\usepackage{amsmath}
\usepackage{amssymb}
\usepackage{booktabs}

\usepackage{amsfonts}
\usepackage{bbding}
\usepackage{caption}
\usepackage{subcaption}
\usepackage{pifont}
\newcommand{\cmark}{\ding{51}}%
\newcommand{\xmark}{\ding{55}}%

\usepackage[pagebackref,breaklinks,colorlinks]{hyperref}
\usepackage{comment}

\usepackage[capitalize]{cleveref}
\crefname{section}{Sec.}{Secs.}
\Crefname{section}{Section}{Sections}
\Crefname{table}{Table}{Tables}
\crefname{table}{Table}{Tables}

\def\cvprPaperID{9772} 
\def\confName{CVPR}
\def\confYear{2022}

\begin{document}

\title{Symmetry-aware Neural Architecture for Embodied Visual Navigation}

\author{Shuang Liu\\
RIKEN Center for AIP\\
{\tt\small shuang.liu.ej@riken.jp}
\and
Okatani Takayuki\\
Tohoku University, RIKEN Center for AIP\\
{\tt\small okatani@vision.is.tohoku.ac.jp }
}
\maketitle

\begin{abstract}
Visual exploration is a task that seeks to visit all the navigable areas of an environment as quickly as possible. The existing methods employ deep reinforcement learning (RL) as the standard tool for the task. However, they tend to be vulnerable to statistical shifts between the training and test data, resulting in poor generalization over novel environments that are out-of-distribution (OOD) from the training data. In this paper, we attempt to improve the generalization ability by utilizing the inductive biases available for the task.  Employing the active neural SLAM (ANS) that learns exploration policies with the advantage actor-critic (A2C) method as the base framework, we first point out that the mappings represented by the actor and the critic should satisfy specific symmetries. We then propose a network design for the actor and the critic to inherently attain these symmetries. Specifically, we use $G$-convolution instead of the standard convolution and insert the semi-global polar pooling (SGPP) layer, which we newly design in this study, in the last section of the critic network. Experimental results show that our method increases area coverage by $8.1 m^2$ when trained on the Gibson dataset and tested on the MP3D dataset, establishing the new state-of-the-art.
\end{abstract}

\section{Introduction}

Embodied visual navigation 
is one of the key problems of autonomous navigation and has attracted increasing attention recently. Researchers have studied various target-oriented tasks of visual navigation so far. These include point goal navigation aiming to reach a given coordinate in an environment \cite{ye2020auxiliary}, object goal navigation aiming to find a specific object in an environment \cite{Chaplot2020ObjectGN,target-driven,gupta2017cognitive}, image goal navigation aiming to navigate to a location in an environment designated by an image \cite{Savinov2018SemiparametricTM,NeuralTopological}, reconstruction navigation deciding where to look next to reconstruct a scene \cite{jayaraman2018learning,seifi2019look}, and room goal navigation aiming to navigate to a specified room \cite{BayesianMemory}. 

In contrast with these tasks, visual exploration is universal and task-agnostic; an agent is asked to visit all the navigable areas of an unknown environment as quickly as possible. Thus, it can serve as an upstream task for the above target-oriented tasks. For instance, the agent may first explore the unseen environment and establish the knowledge about the environment, after which the agent utilizes it to perform specific tasks efficiently \cite{Savinov2018SemiparametricTM}.

Numerous methods have been proposed for the problem of visual exploration so far. These methods are divided into two categories, heuristic methods \cite{frontierexploration,Fastfrontier,beyondfrontier} and learning-based methods. 
Recent studies have mainly considered learning-based methods. 

Visual exploration is a problem of choosing the optimal action at each time step to maximize the above exploration goal, given a history of inputs to the agent. Recent studies employ simulators of virtual environments, such as Habitat \cite{szot2021habitat}, to consider the problem under realistic conditions. In such simulators, photorealistic visual inputs from virtual environments are available. To predict the optimal action from the agents’ input, including the visual inputs, learning-based methods typically use reinforcement learning (RL) to acquire the mapping between the two from an enormous amount of agents’ interaction with the virtual environments. 

Although they have achieved reasonably good exploration performance, the learning-based methods are inherently vulnerable to the statistical deviations between training and test data. Specifically, a model trained on a set of training data works well on novel environments that are statistically similar to the training data but yields suboptimal performance on novel environments that are dissimilar, i.e., out-of-distribution (OOD) inputs. This is not the case with the heuristic methods. 

To apply the learned models to real-world environments, it is necessary to increase their generalization ability in the above sense. How is it possible? A promising approach is data augmentation, and it has been widely adopted  in both deep learning \cite{calimeri2017biomedical,madani2018chest,t2fgan} and reinforcement learning \cite{raileanu2021automatic,laskin2020reinforcement,Kostrikov2021ImageAI}. However, this approach has its limitations.

In this study, we pay attention to the inductive biases specific to the task. Using them, we want to narrow the solution space for RL and thus avoid overfitting to training data, aiming to improve the above generalization ability. Specifically, we employ the framework of active neural SLAM (ANS) \cite{ANS}, which is the most successful for the task, and consider how to do the above within the framework. 

ANS is a method with a modular and hierarchical structure. First, a module yields a long-term goal from a history of the agent’s inputs. The mapping between them is called global policy, which is represented by a neural network. Next, another module computes the shortest path from the agent’s position to the given long-term goal. The path is then subsampled to construct a sequence of short-term goals. Selecting a short-term goal in turn, another module generates actions to reach the short-term goal.

The module that plays a key role in ANS is the neural network that represents the global policy. (The problems solved by the other modules are not so hard.) ANS employs the advantage actor-critic (A2C) method to learn the global policy. The actor and the critic are modeled by neural networks, specifically, a pair of independent networks or a single network having two output branches. The actor and the critic receive as their input the current estimate of the environment’s map along with the agent’s navigation paths etc. expressed in the form of a multi-channel 2D map. The actor yields a 2D likelihood map for the long-term goal, whereas the critic predicts the future accumulated reward as a scalar value. 

In this paper, we first point out that specific symmetries should exist for each mapping represented by the actor and the critic. Concretely, the 2D map predicted by the actor should be equivariant to translation and rotation of its input having the form of a 2D map. The reward predicted by the critic should be invariant to input rotation and should not be invariant to input translation. However, standard convolutional neural networks (CNNs), employed in ANS, do not by nature have these symmetries. 


To cope with this, we design the network to inherently possess the desired symmetries for the actor and the critic. Specifically, we redesign the network of ANS as follows. First, we employ $p4$ $G$-convolution \cite{pmlr-v48-cohenc16}, which approximately achieves rotation equivariance, for all the convolutional layers and replace the max pooling with blur pooling \cite{zhang2019making}, which achieves translation equivariance more accurately. Then, the resulting network representing the actor becomes equivariant to input translation and approximately to input rotation. For the critic, employing the same stack of convolutional layers (rigorously, it is shared by the actor), we design and place a new building block named a semi-global polar pooling (SGPP) layer on top of the convolutional block. The SGPP layer applies polar mapping to its input and then pooling in the circumferential direction in the input map. Then, the resulting critic becomes not invariant to input translation and approximately invariant to input rotation.

We conduct experiments to test our approach using the Gibson \cite{xiazamirhe2018gibsonenv} and the MP3D \cite{Matterport3D} datasets. The results show that the proposed method improves the original ANS by a good margin, proving the approach's effectiveness. Specifically, it improves ANS by 8.1 $m^2$ (from 76.3 to 84.4 $m^2$) in area coverage, the primary metric for evaluating exploration performance, in the setting to evaluate the generalization ability, i.e., training the models on Gibson and testing them on MP3D.

\section{Related Works}

\subsection{Visual Navigation}

Classical navigation methods leverage simultaneous localization and mapping (SLAM) to construct environmental maps, including occupancy map \cite{cadena2016past} and topological map \cite{Savinov2018SemiparametricTM}. Agents navigate to points in the environments. This is known as map-based navigation \cite{bonin2008visual}. 

In contrast with the map-based navigation that assumes a known map, target-driven navigation tasks have recently emerged, which are termed mapless navigation. In these tasks, the movement of the agent is determined by visual cues provided by the environment. Example tasks are object-goal, room-goal, and image-goal navigation. 

Object-goal navigation requires agents to navigate to designated objects in an unknown environment. Either implicitly learned \cite{target-driven} or explicitly encoded \cite{du2020learning,lv2020improving} relationship of objects is utilized to facilitate finding the objects. Room-goal navigation requires agents to arrive at a specific room as quickly as possible. A method is proposed in \cite{BayesianMemory} to learn a probabilistic relation graph to acquire prior knowledge about the layout of environments. 

Image-goal navigation asks agents to navigate to a location in an unknown environment that is specified by its image. A popular solution is to take both current and goal observation as inputs and employ a Siamese network to perform the navigation effectively. The method proposed in \cite{mezghani2021memory} combines an attention mechanism with the Siamese network for building the memory of environments, which is then
used by the policy for image-driven navigation. The work \cite{choi2021image} follows the same formulation based on a Siamese network but exploits information obtained through keypoint matching, generating a self-supervised reward.

\subsection{Visual Exploration}

Visual exploration has received considerable attention due to its task-agnostic nature \cite{jayaraman2018learning,mezghani2020learning,ANS,ramakrishnan2020occupancy,Chen2019LearningEP}. Classical approaches continually select vantage points, such as frontier points \cite{frontierexploration}, to visit. The downside of these approaches is that they are vulnerable to external noises such as sensor noises.

The recent boom of deep reinforcement learning (RL) has created a new wave of development of visual exploration. Numerous works have casted the visual exploration as a partially observable Markov decision process and solve it under the framework of RL \cite{jayaraman2018learning,mezghani2020learning,beeching2020egomap,nagarajan2020learning,pathak2017curiosity,Chen2019LearningEP,ANS,ramakrishnan2020occupancy,ramakrishnan2021exploration,qi2020learning}. Most of them use an actor-critic architecture \cite{beeching2020egomap,nagarajan2020learning,pathak2017curiosity,ANS,ramakrishnan2020occupancy,ramakrishnan2021exploration}. 

The actor learns a navigation policy, and the critic estimates a value function given the agent's current state. Existing approaches to improve performance have been geared towards novel designs of reward, such as coverage, curiosity, and novelty \cite{ramakrishnan2021exploration}, or novel architecture design, e.g., the hierarchical architecture \cite{ANS}. There is a study \cite{Chen2019LearningEP} that proposes to utilize area coverage reward along with supervision from human experience to learn policies. 

An excellent standard framework, which is actor-critic based as well, was constructed by \cite{ramakrishnan2021exploration} to evaluate multiple rewards for exploration. The approaches above all solve the exploration problem in an end-to-end manner, directly mapping visual data to actions. 

In contrast to this, the work \cite{ANS} tackled it hierarchically. It divided the mapping into two steps. First, a global target point, which was similar to vantage points in classical exploration methods, was computed by a global policy network. Then a local policy network produces actions to reach the waypoints generated by the path planner according to the global target point. Occupancy anticipation \cite{ramakrishnan2020occupancy} also predicts and rewards invisible spatial maps to improve map accuracy. 

So far, existing methods have produced impressive results in the case where the testing environments are similar to training environments in terms of layout, area, etc. Nevertheless, they still struggle with the cases where training and testing environments are different, hindering their deployment in practice. It is primarily because RL notoriously struggles with (poor data efficiency and) generalization capabilities. We aim to enhance generalization ability for visual navigation by injecting inductive bias about symmetry into networks.  

\subsection{Equivariance and Invariance}

Recent years have witnessed the great success of convolutional neural networks (CNNs) in computer vision tasks. CNNs have built-in translation-equivariance, and those with downsampling/pooling operations have built-in local translation invariance. These contribute to the successes in applications to various problems. It could be possible to learn these symmetries from data if the model has sufficient parameters. However,  it will increase the risk of overfitting. 

A function is equivariant if the output changes in the same way that the input changes \cite{Goodfellow-et-al-2016}. Novel convolution layers with different equivariance have been proposed so far, group equivariant convolution networks \cite{pmlr-v48-cohenc16,ICML-2016-DielemanFK}, steerable convolution networks and harmonic networks for rotation equivariance, scale equivariance \cite{lindeberg2021scale,worrall2019deep}, and permutation equivariance \cite{thiede2020the}. Such convolution layers equipped with various types of equivariance have been proven to be beneficial to better performance 
in tracking \cite{sosnovik2021scale}, classification, trajectory prediction \cite{walters2020trajectory}, segmentation \cite{muller2021rotation}, and image generation \cite{dey2020group}. However, these methods have never been applied to visual navigation. 

A primary purpose of pursuing various types of equivariance is to achieve a certain type of invariance. A mapping is called invariant if the output remains the same no matter how the input changes. Although data augmentation and invariance-oriented loss functions \cite{cheng2016learning} may enhance global invariance, it is not guaranteed to generalize to out-of-distribution data. By contrast, global invariance can be imposed by a global pooling layer following equivariant layers. Global rotation invariance is enforced in \cite{cheng2016learning} for texture classification by combining equivariant convolution layers with a global average pooling layer in this order. Several studies \cite{lindeberg2021scale,sosnovik2019scale} attain global scale-invariance by combining layers having scale-equivariance with global max pooling. 


\section{Symmetry-aware Neural Architecture}

\subsection{Problem of Visual Exploration}

Following previous studies \cite{ANS,Chen2019LearningEP,ramakrishnan2020occupancy}, we consider a visual exploration task in which an agent explores an unknown 3D environment, e.g., a floor of a building, that allows only 2D motion. The goal for the agent is to go everywhere in the environment it can go while creating an environment's 2D map. We employ the Habitat simulator \cite{habitat19iccv} as the framework for studying the task. 

The agent receives several inputs from the environment, i.e., an image or a depth map of the scene in front of it, the agent's pose, and an odometry signal measuring its motion. Then, the agent moves to explore the environment by feeding the actuation signal to its motor. Thus, the problem is to compute the actuation signal and update the environment map at every time step, given the history of the inputs the agent has received until then. See the literature \cite{ANS} for more details. 

Recent studies have formulated the problem as learning policies that yield actions (or intermediate representation leading to actions) from the received inputs; most of them employ reinforcement learning (RL) 
\cite{Chen2019LearningEP,ANS,ramakrishnan2020occupancy,jayaraman2018learning}. 
Recently, Chaprot et al. \cite{ANS} proposed a method named active neural SLAM (ANS), having established the new state-of-the-art. Our study is built upon ANS and we will first revisit and summarize it below.


\subsection{Reivisiting Active Neural SLAM (ANS) }

ANS has a modular and hierarchical structure to better solve the visual exploration problem. Instead of learning a direct mapping from the inputs to an action, ANS learns two different policies in a hierarchy, i.e., a global policy that yields a long-term goal in the 2D environment map and a local policy that yields actions to approach a short-term goal, which is subsampled from the path to the long-term goal.
More specifically, ANS consists of four modules, i.e., a neural SLAM module, a global policy module, a path planner, and a local policy module. 

The neural SLAM module computes a local top-view 2D egocentric map $p_t\in[0,1]^{2\times v \times v}$ and estimates an accurate agent pose from the current inputs. (The first and second channel of $p_t$ represents the obstacle and the explored region at time $t$, respectively.) Then, the computed local egocentric map is registered to the global map $h_t \in [0,1]^{4 \times M \times M}$ using the estimated pose. (The first and second channel of $h_t$ represents the obstacles and explored area at time $t$ respectively. The third channel indicates the computed agent position. The last channel records the path that has been visited by the agent.)

The global policy module makes ANS the most distinct from other methods. It predicts a long-term goal $g_t$ given two different views $h_t^g$ and $h_t^l$ 
of the latest map $h_t$; $h_t^g$ is a rescaled version of $h_t$, and $h_t^l$ is a local view cropped from $h_t$ with the agent position as its center; and $h_t^g$ and $h_t^l$ have the same size $G\times G$ and are concatenated in the channel dimension.

The module is implemented as an advantage actor-critic (A2C) network \cite{mnih2016asynchronous}. As shown in \cref{fig:netarc}, it consists of an actor network and a critic network. Each network has a similar design of a convolutional block, which consists of a stack of convolutional and max-pooling layers, and subsequent fully connected layers. In a standard design, they share the convolutional block. 

The actor network outputs a 2D likelihood map of the long-term goal $g_t$ represented in the coordinates of the global map $h^g_t$. It learns a global policy, denoted by $g_t=\pi(s_t|\theta_G)$, where $s_t={(h_t^l,h_t^g)}$ is the state at time $t$ and $\theta_G$ are parameters. The critic network represents a value function $V(s_t|\theta_V)$; $V(s_t|\theta_V)$ estimates the expected future accumulated reward with the agent, which is currently at state $s_t$ and will take actions by following the current policy $\pi(s_t|\theta_G)$. The parameters $\theta_G$ and $\theta_V$ are optimized to maximize the area coverage. 



The other two modules, i.e., the path planner and the local policy module, play the following roles. The path planner computes the shortest path from the current location to the long-term goal, which is subsampled to  generate a number of short-term goals. The local policy module predicts actions to reach the next short-term goal.

\subsection{Symmetries in the Global Policy}

\subsubsection{Outline}

Although ANS performs reasonably well and is currently state-of-the-art, its network is a plain CNN consisting only of generic network components, such as convolutional and fully connected layers. Thus, there is room for improvement in architectural design, and we aim to find a design that better fits the problem structure. 

Toward this end, sticking to the basic framework of ANS that is proven to be effective, we propose to redesign its global policy module, which consists of the actor and critic networks. We consider what conditions the mappings realized by the two networks should satisfy. We pay attention to symmetry of the mappings, more specifically, their equivariance and invariance to translation, rotation, and scaling. Leveraging the fact that the exploration task is a geometric problem, we derive what symmetry the ideal actor and critic networks should implement, based on which we redesign the two networks.

In other words, we incorporate the inductive bias of the task into the network design. A neural network represents a mapping, and we train it using data to hopefully find an optimal mapping. The network’s architecture specifies a subspace in the mapping space. Enforcing the network to inherently satisfy a required constraint by its design shrinks the subspace, leading to a better solution. 


\subsubsection{Invariance and Equivariance}

As mentioned above, we are interested in \textit{equivariance} and \textit{invariance}. Their mathematical definitions are as follows. 
Given a group $\mathcal{G}$ on a homogeneous space $ \mathcal{X}$, a mapping $\Psi: f(x)  \rightarrow f'(x)$ that has the property 
\begin{equation}
\Psi[\mathbb{T}_g f(x)]=\mathbb{T}_g'\Psi [f(x)], \forall f, x \in \mathcal{X},g \in \mathcal{G},
\end{equation}
is said to be {\em equivariant} to the group $\mathcal{G} $ (or $\mathcal{G}$-equivariant) if $g=g'$ and to be {\em invariant} to the group $\mathcal{G}$ (or $\mathcal{G}$-invariant) if $g=g'=e$, the identity. $\mathbb{T}_g$ is the transformation corresponding to its group action $g$. 

Intuitively, they are intepreted as follows. Suppose a mapping that receives an input and yields an output and also applying a geometric transformation to the input of the mapping. If the mapping is invariant to the transformation, the output for the transformed input will remain the same as the original output. If the mappping is equivariant to the transformation, that will be the same as the result of applying the same transformation to the original output. 


Convolutional layer are equivariant to input translation. Thus, shifting an input map results in the same shift of the output map. This applies to the input/output of a stack of any number of convolutional layers. (Rigorously, the output map undergoes the input shift plus downscaling corresponding to the downsampling in those layers.) On the other hand,  global average pooling (GAP) layers are invariant to input translation. Thus, CNNs having a GAP layer on top of the stack of convolution layers are invariant to input translation. CNNs and a stack of convolution layers are not equivariant to rotation nor scaling. They are not invariant to them, either. 



\subsubsection{Ideal Symmetries of Actor and Critic Networks}

 
Now we consider what symmetry the actor and critic networks should have. First, the actor network receives a 2D map encoding the current state of the agent,
and outputs a 2D map containing a long-term goal, as explained above and shown in \cref{fig:netarc}.
It is easy to see that this mapping should ideally be equivariant to translation and rotation. It 
should not be equivariant to scaling. 


The critic network receives the same input map and outputs a scalar value, the future accumulated reward when following the current policy given the current state. The mapping should be invariant to rotation since the future accumulated reward should be independent of the orientation of the input map. It should not be invariant to translation or scaling, since applying these transformation to the input map should change the reward.

\cref{tb:symana} (the first two rows) summarizes the requirements for the ideal actor and critic networks. 

To design a new critic network, we limit our attention to networks having a stack of convolutional layers at their initial section. This is also the case with the critic network of ANS, which has a stack of fully connected layers right after the convolutional layer stack, and we will redesign this section. Now, for the mapping from the input to the final output to be invariant to rotation as mentioned above, the mapping to the last convolutional layer needs to be equivariant to rotation. This is because otherwise, it will be extremely hard, if not impossible, to achieve the rotation-invariance at the final output. We express this requirement in the `$\uparrow$ at the last conv.’ row in \cref{tb:symana}.

The actor and critic networks of the original ANS does not have the same symmetries as the ideal ones. The missing symmetries are indicated by the {\color{red} \xmark}'s in \cref{tb:symana}. Specifically, the actor should be rotation-equivariant but is not; the critic should be rotation-invariant but is not. As discussed above, the mapping from the input to the last convolutional layer of the critic should be rotation-equvariant but is not. 

\begin{table}[]
\centering \footnotesize
\begin{tabular}{l|ccc|ccc}
\toprule
         & \multicolumn{3}{c|}{Equivariance} & \multicolumn{3}{c}{Invariance} \\
         & Trans.\!\!      & Rot.\!\!     & Scale & Trans.\!\! & Rot.\!\!   & Scale \!\! \\
\midrule
{Ideal actor}  &  \cmark      &\cmark     & \xmark    &    &  \\

{Ideal critic} &        &  &        &  \xmark &\cmark &\xmark     \\ 
\textit{$\uparrow$ at the last conv.} &  --       &\cmark       & -- &     & &     \\ 
\hline
ANS actor  &  \cmark      &\textcolor{red}{\xmark}     & \xmark\\
ANS critic &     & &     &\xmark      &\textcolor{red}{\xmark}    &\xmark     \\
\textit{$\uparrow$ at the last conv.} &  \cmark         &{\color{red}\xmark} & \xmark&     & &     \\ 
\bottomrule
\end{tabular}
\caption{Upper three rows: ideal symmetries that should be implemented by the actor and critic networks of the ANS framework. Lower three rows: actual symmetries realized by the original actor and critic networks of ANS. \cmark\, indicates the network is equipped with the symmetry. \xmark\,  indicates the network does not have the symmetry. `--' indicates the symmetry is not specified. Blank cells mean the symmetry is irrelevant.  {\color{red}\xmark}\, in red color indicates that the symmetry of ANS networks differs from the ideal one. }
\label{tb:symana}
\end{table}

\subsection{Rotation-Equivariance of the Actor Network}

Thus, we need to newly equip the actor network with rotation-equivariance while maintaining its translation-equivariance. To do so, we propose to employ \textit{G}-convolution \cite{pmlr-v48-cohenc16} and blur pooling \cite{zhang2019making}; specifically, we replace the standard convolution and max pooling of the original ANS actor network with the two, respectively.




Considering computational efficiency, we employ $p4$ \textit{G}-convolution. 
All combinations of translations and rotations by $90$ degrees form the group $p4$, which can be parameterized by 
\begin{equation}
g(m,z_1,z_2) = 
    \begin{bmatrix}
    \cos(m\pi/2) & -\sin(m\pi/2) & z_1 \\
    \sin(m\pi/2) & \cos(m\pi/2)  & z_2 \\
    0           & 0         &  1
    \end{bmatrix},
\end{equation}
where $m\in\{0,1,2,3\}$ and $(z_1,z_2) \in \mathbb{Z}^2$ \cite{pmlr-v48-cohenc16}.

\begin{figure}
\centering
\includegraphics[width=0.5\textwidth]{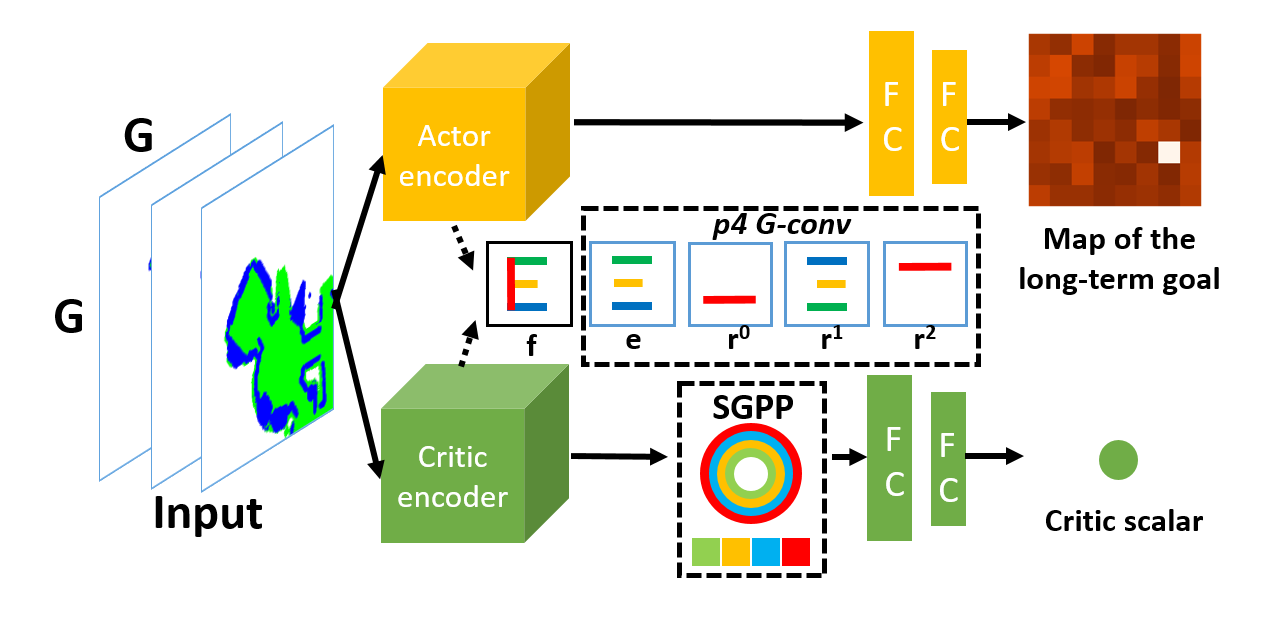}
\caption{Overall architecture of Symmetric Global Policy Network (using \textit{G}-convolution, blur pooling and SGPP) and Global Policy Network of ANS (using convolution, max pooling and without using SGPP). The red dashed circle illustrates the feature maps produced by $p4$ \textit{G}-convolution assuming its input is \textbf{E} and the filter is a horizontal edge extractor.}
\label{fig:netarc}
\end{figure}

As shown in \cref{fig:netarc}, 
$p4$ \textit{G}-convolution can be performed by first rotating filters 
with angles $m\pi/2, m =\{0,1,2,3\}$ to form a filter bank, and then applying it to 
the input feature map.  

As mentioned above, we redesign the convolutional layer stack of the actor network with the same number of layers performing $p4$ \textit{G}-convolution. The original ANS actor network employs max pooling for downsampling the feature map. However, max pooling makes the translation-equivariance inaccurate. Max pooling ignores the Nyquist sampling thereom, breaking translation equivariance. Blur pooling filters the signal before downsampling to better preserve translation equivariance. Hence, we employ blur pooling and replace max pooling with this.  




By revising the convolutional block (i.e., the stack of convolutional and pooling layers) in the original ANS actor network as above, the mapping represented by the block becomes approximately equivariant to rotation and precisely equivariant to translation. We need to maintain the fully connected layers after the convolutoinal block since predicting the agent's goal seems to need them to integrate the global features in a non-simple manner. Fully connected layers are not inherently equipped with the desired equivariance, and we leave it to training; owing to their flexibility, we expect the actor network to gain the desired equivariance at the final output.

On the other hand, we want to make the critic network have rotation-invariance. As mentioned earlier, to do so, we want the mapping represented from the input to the output of the convolutional block to be rotation-equivariant. We employ here the design of the original ANS that the critic network share the convolutional block with the actor network. Then, the above redesign of the convolutional block achieves what we want. We further revise the section on top of the convolutional block to realize the desired symmetries for the critic network. 

\subsection{Rotation-Invariance of the Critic Network: Semi-global Polar Pooling}

The critic network should represent a mapping that is rotation-invariant and is not translation-invariant or scale-invariant, as shown in \cref{tb:symana}. We assume the convolutional block to have rotation-equivariance due to its design, as explained above. To attain the above (in)variance, we then propose a new network component, named the semi-global polar pooling (SGPP) layer.  

\begin{figure}
\centering
\includegraphics[width=0.35\textwidth]{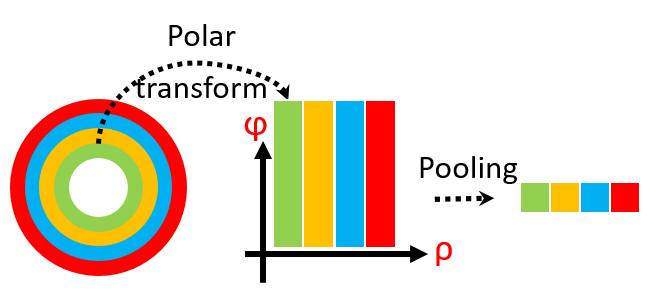}
\caption{Illustration of the proposed semi-global polar pooling (SGPP).}
\label{fig:SGPP}
\end{figure}

SGPP is illustrated in \cref{fig:SGPP}. It first converts the input feature map from Cartesian space to polar space. Let $I(x,y) \in \mathbb{R}^{c\times h \times w}$ denote the feature map represented in Cartesian space, where $x,y$ are the Cartesian coordinates. The feature map $I'(\rho,\phi) \in \mathbb{R}^{c\times h \times w}$ represented in polar space is given by 
\begin{equation}
\begin{aligned}
\rho  & = \sqrt{x^2+y^2},  \\
\phi &= \arctan \frac{y}{x},
\end{aligned}
\end{equation}
where $\rho$ and $\phi$ are the coordinates in polar space. We finally apply average-pooling to the features over all $\phi$'s as
\begin{equation}
\Phi = \frac{1}{n}\sum_\phi I'(\rho,\phi),
\end{equation}
to obtain a pooled feature $\Phi \in \mathbb{R}^{c\times h}$. 

As the convolutional block, whose output is the input to the SGPP layer, is equivariant to rotation, $\Phi$ is invariant to rotation since it is pooled over the coordinate $\phi$ (i.e., the circumferential direction). $\Phi$ is further processed by fully connected layers, computing the final output of the critic network. As $\Phi$ is already invariant to rotation, these additional layers do not change the invariance. While the convolutional block is also equivariant to translation, the Cartesian-polar conversion invalidates the equivariance; thus $\Phi$ is not invariant to translation. It is not invariant to scaling, either. Thus, the critic network having the SGPP layer in between the convolutional block and the fully connected layers attains the desired (in)variance to translation, rotation, and scaling, as illustrated in \cref{tb:symana}.

\section{Experimental Results}

\subsection{Experimental Setup}

As with previous studies \cite{NeuralTopological,ANS,ramakrishnan2020occupancy,Chen2019LearningEP,Chaplot2020ObjectGN,gan2020look,chaplot2020semantic}, we use the Habitat simulator \cite{habitat19iccv} for our experiments. We choose the configuration that depth images are available as the visual input and the actuation and sensory signals include noises. This applies to the training and test times. To evaluate the performance of exploration methods, we use the area coverage (i.e., the area seen during exploration) within a time budget for the primary metric, following\cite{Chen2019LearningEP,ANS}. We compare our method with several baselines. We run each method five times and report their average area coverage with the standard deviation. Our code is included in the supplementary material.


\subsection{Datasets}

\begin{figure}
\centering
  \begin{subfigure}{\linewidth}
    \includegraphics[width=\linewidth]{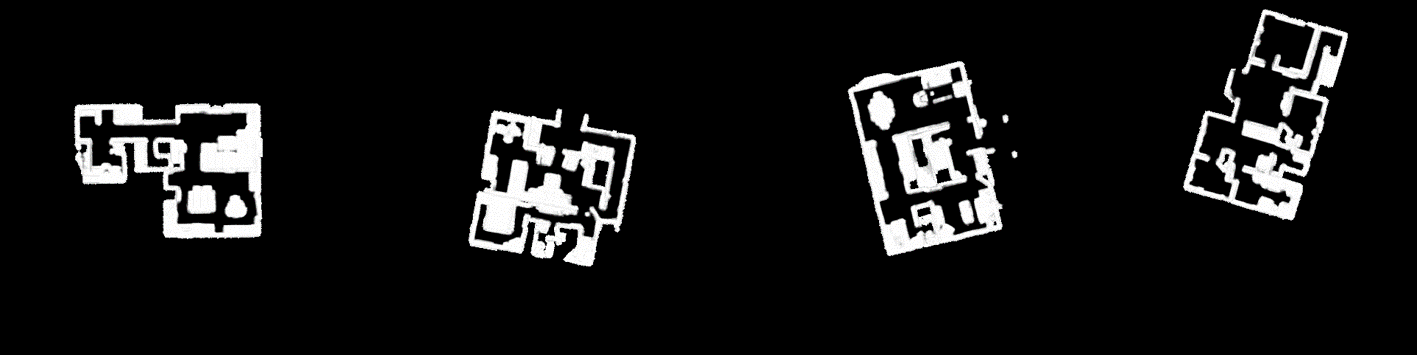}
    \caption{Gibson.}
    \label{fig:sample-gibson}
  \end{subfigure}
  \begin{subfigure}{\linewidth}
    \includegraphics[width=\linewidth]{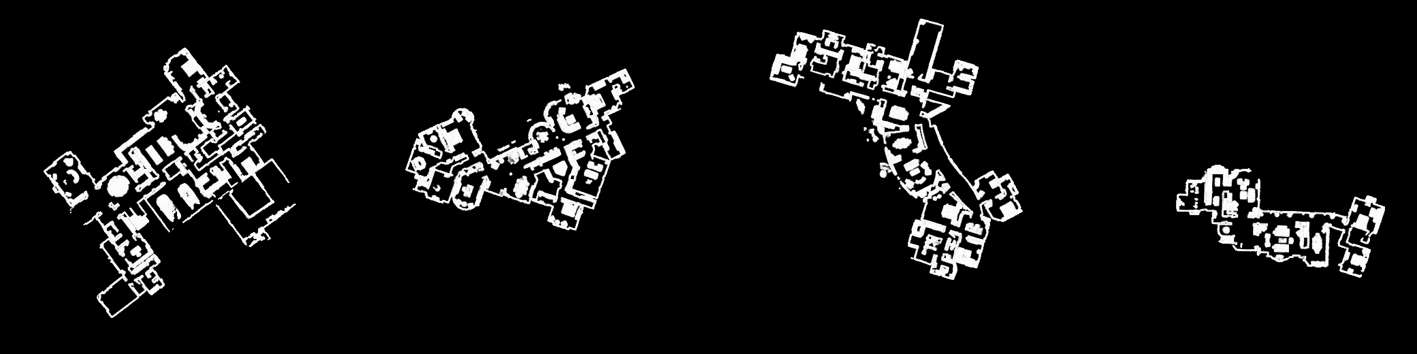}
    \caption{MP3D.}
    \label{fig:sample-mp3d}
  \end{subfigure}
  \caption{Environment layout of samples from (a) Gibson and (b) MP3D dataset.}
  \label{fig:datasample}
  \end{figure}

Following previous studies, we employ two datasets, Gibson \cite{xiazamirhe2018gibsonenv} and MP3D \cite{Matterport3D}. Both of them contain photorealistic virtual environments created from real-world scenes. Most environments contained in Gibson are office spaces, while those in MP3D are homes. Thus, the environments of Gibson differ from those of MP3D in terms of scenarios, layout, area, and so on. Overall, MP3D environments are larger in the area and more irregular in the layout. \cref{fig:datasample} provides the layouts of several representative examples from Gibson and MP3D. It is seen that there is a gap in various aspects between the two. 

To evaluate the generalization ability of the methods across different datasets, we train each model on Gibson and test it either on Gibson (in-distribution) or on MP3D (out-of-distribution), following the experiments of \cite{ANS}.


\subsection{Compared Methods}

We experimentally compare the proposed method with several baselines for global policy.
The first is FBE (Frontier based Exploration)\cite{frontierexploration}. FBE is a heuristic method that iteratively selects a point at the frontiers (boundaries between explored free region and unexplored region) using various strategies. We follow the strategy implemented in \cite{ramakrishnan2021exploration}, selecting a random point at the longest boundary. We replace the global policy module in ANS with FBE and evaluate its performance. 

The second is a variant of FBE, named FBE-RL, which replaces the global policy module in ANS with a RL-based FBE. To be specific, FBE-RL selects a target point at the frontiers using the policy produced by an actor-critic network, instead of a heuristic policy. More details of FBE-RL will be given in the supplementary material. 

The third is ANS, for which we use the code\footnote{github.com/facebookresearch/OccupancyAnticipation} and the settings given in \cite{ramakrishnan2020occupancy}. Our method for generating global policy is integrated to ANS by replacing the original global policy module with ours. We call this model symmetry-aware ANS or S-ANS in what follows. We run each model five times and report their average and  standard deviation of the area coverage at time step $1,000$. 


\subsection{Results}

\cref{tab:area} shows the results. Comparing the methods in the case of training and testing on Gibson, the proposed S-ANS outperforms ANS with the margin of $0.8m^2$ ($33.7$ vs. $32.9m^2$). The margin becomes more significant when testing the same models on MP3D, i.e., $8.1m^2$ ($84.4$ vs. $76.3m^2$). This demonstrates the better generalization ability of S-ANS. 

\begin{table}[t]
\centering
\begin{tabular}{ccc}
\toprule
            & Gibson ($m^2$)  & MP3D ($m^2$)   \\
\midrule
FBE         & $26.5 \pm 0.5$              & $69.6 \pm 1.9$   \\
FBE-RL      & $28.2  \pm 0.3$              & $63.0 \pm 3.2$   \\
ANS         & $32.9 \pm 0.2 $ & $76.3 \pm 2.8$ \\
S-ANS       & $\mathbf{33.7 \pm 0.2}$ & $\mathbf{84.4 \pm 1.7}$ \\
\bottomrule
\end{tabular}
\caption{Exploration performance (in area coverage, $m^2$) of different models on Gibson and MP3D. All the  models are trained on Gibson.}
\label{tab:area}
\end{table}

\begin{figure}[t]
\centering
  \begin{subfigure}{\linewidth}
    \includegraphics[width=\linewidth]{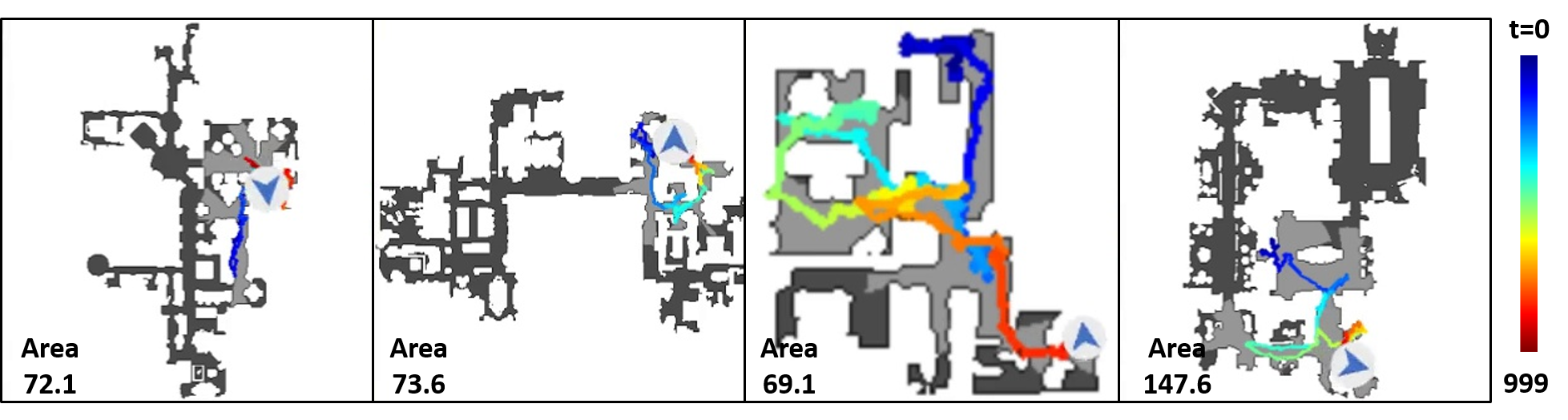}
    \caption{ANS.}
    \label{fig:path-ans}
  \end{subfigure}
  \begin{subfigure}{\linewidth}
    \includegraphics[width=\linewidth]{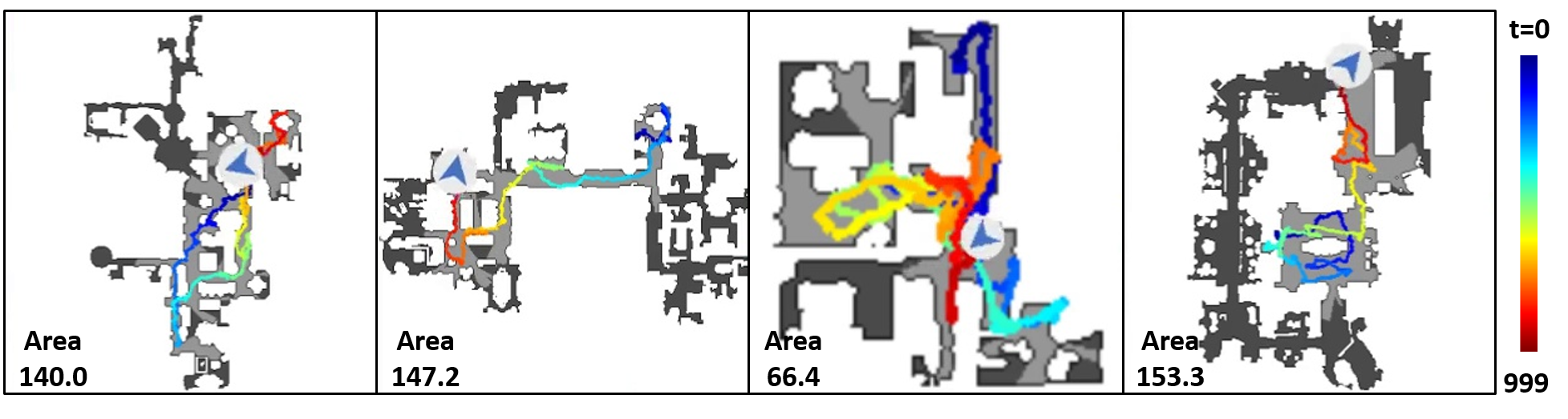}
    \caption{S-ANS.}
    \label{fig:path-sans}
  \end{subfigure}
  \caption{Exploration paths of ANS (top row) and S-ANS (bottom row) for four representative 
  environments from MP3D. 
    The number at the bottom left of each box indicates the area coverage ($m^2$). The dissimilarity of the environment from the training data tends to widen the gap between the two methods.  }
  \label{fig:path}
  \end{figure}

\cref{fig:path} shows the representative examples of their exploration paths of ANS and S-ANS on the same four environments of MP3D. The first and second columns of \cref{fig:path} show typical examples for which S-ANS explores a much larger area than ANS; S-ANS explores twice as large area as ANS. As with these examples, S-ANS tends to show better exploration performance for environments that are more dissimilar from those of Gibson. When the environments are similar, their performances tend to be close, as shown in the third and fourth columns of \cref{fig:path}. These observations further verify the improved generalization ability of S-ANS. 

It is seen from \cref{tab:area} that FBE and FBE-RL show inferior performance. It is noteworthy that FBE and FBE-RL behave differently for different combinations of the train and test data. While FBE-RL performs better on Gibson by $1.7m^2$ ($28.2$ vs $26.5m^2$), it performs worse on MP3D by $-6.6m^2$ (63.0 vs $69.6m^2$). We can say that learning better policies improves exploration performance when there is only a little gap between the train and test data. However, it is a double-edged sword; it causes overfitting and leads to inferior exploration ability when there is a gap between the train and test data. This demonstrates the difficulty with learning-based approach to the exploration task. 



\subsection{Ablation Study}

As explained earlier, our method imposes translation- and rotation-equivariance on the actor network and (approximate) rotation-invariance on the critic network. The former is implemented by the \textit{G}-convolution and the blur pooling. The latter is implemented by SGPP in addition to \textit{G}-convolution and the blur pooling. To examine the effectiveness of each component, we create two variants of the proposed network, named E-ANS and G-ANS. 

E-ANS is a model created by removing the SGPP layer from the critic network of S-ANS. Without SGPP, it is not (even approximately) invariant to rotation, translation, or scaling. Its actor network maintains the ideal property of equivariance. G-ANS is an intermediate model between E-ANS and S-ANS; it is created by replacing the SGPP layer with a global average pooling (GAP) layer in the critic network of S-ANS. The added GAP layer makes the critic network invariant to both rotation and translation; recall that the convolutional block before the GAP layer is equivariant to rotation and translation due to the employment of the $G$-convolution and the blur-pooling. The added translation invariance is superfluous compared with the ideal invariance of the critic network shown in \cref{tb:symana}. 
In short, considering the excesses and deficiencies of the implemented invariance/equivariance, the expected performance will be S-ANS $>$ E-ANS, G-ANS $>$ ANS.



\begin{figure}
\centering
\includegraphics[width=0.7\linewidth]{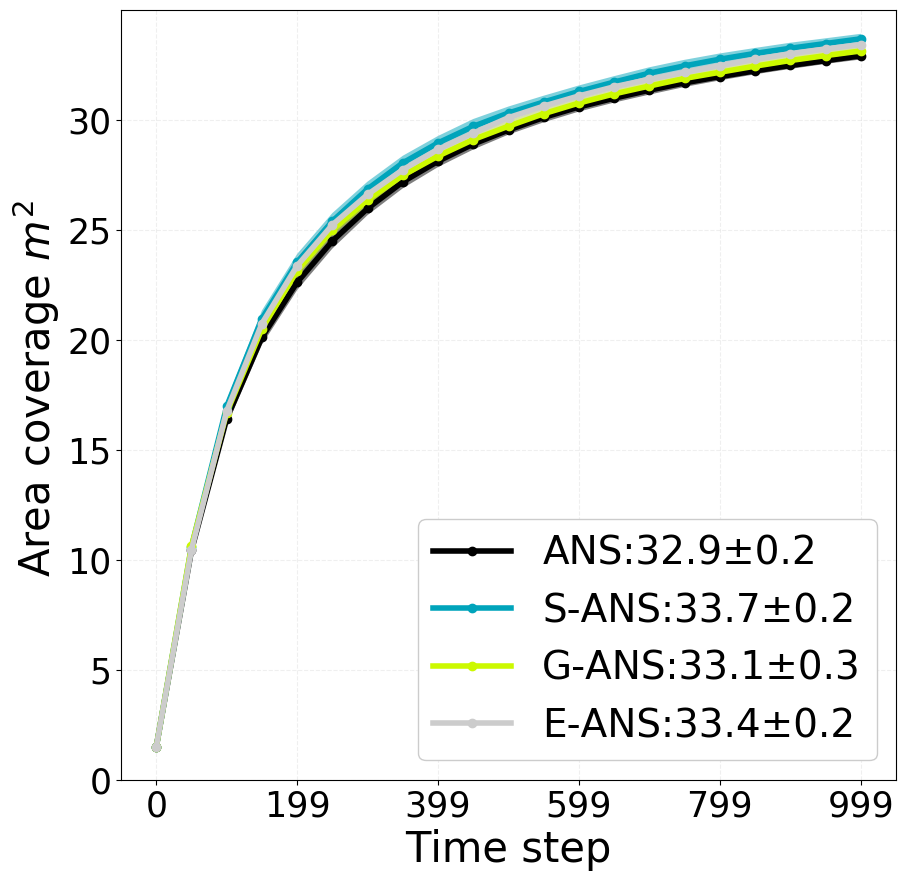}\\
\raisebox{3mm}{\small (a) Gibson}\\

\includegraphics[width=0.7\linewidth]{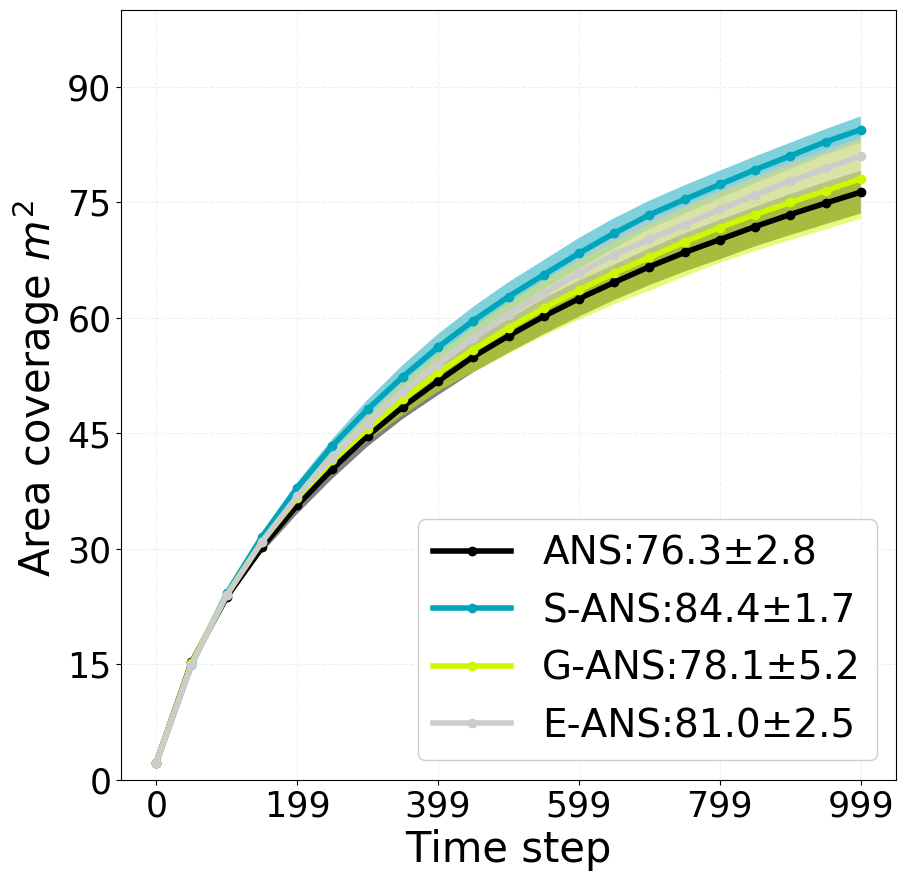}\\  
\raisebox{3mm}{\small (b) MP3D}
\vspace*{-3mm}
  \caption{Expolaration performance (in area coverage, $m^2$) of ANS, S-ANS, G-ANS, and E-ANS tested on (a) Gibson and (b) MP3D. All the models are trained on the Gibson dataset. }
  \label{fig:ablation}
  \end{figure}

We train E-ANS and G-ANS on Gibson and test them on Gibson and MP3D in the same way as ANS and S-ANS. Figs.~\ref{fig:ablation}(a) and (b) show the explored areas by the four models on the test splits of Gibson and MP3D, respectively. Each solid curve and shadowed area indicates the mean and standard deviation over five runs, respectively, of the area coverage at a different time step.  

It is seen from the results that the four models are ranked in the performance as S-ANS $>$ E-ANS $>$ G-ANS $>$ ANS. This matches well with our expectation mentioned above. The results further tell us that the excessive translation-invariance of the critic network in G-ANS does more harm than deficient rot-invariance of that of E-ANS. Overall, these results validate the effectiveness of our approach. 

\section{Conclusion}
Existing learning-based methods for visual exploration struggle with generalization to OOD environments, that is, statistically different environments from those used for training. We believe this is partly because the inductive biases available for the task are not used effectively. Employing the framework of active neural SLAM (ANS), we have shown that the actor and the critic should satisfy specific symmetries with their mappings. We then propose a design of neural networks that inherently possesses the ideal symmetries. Specifically, we propose to use \textit{G}-convolution instead of the standard convolution layer.  We also propose the semi-global polar pooling (SGPP) layer, a network component, that can make the network invariant to rotation and not invariant to translation when using it at the final section of the critic network. Our approach improves the performance of ANS by $8.1m^2$ in area coverage in the setting where training and test datasets are different.
{\small
\bibliographystyle{ieee_fullname}
\bibliography{ref}

\begin{thebibliography}{10}\itemsep=-1pt

\bibitem{beeching2020egomap}
Edward Beeching, Jilles Dibangoye, Olivier Simonin, and Christian Wolf.
\newblock Egomap: Projective mapping and structured egocentric memory for deep
  rl.
\newblock In {\em Joint European Conference on Machine Learning and Knowledge
  Discovery in Databases}, pages 525--540. Springer, 2020.

\bibitem{bonin2008visual}
Francisco Bonin-Font, Alberto Ortiz, and Gabriel Oliver.
\newblock Visual navigation for mobile robots: A survey.
\newblock {\em Journal of intelligent and robotic systems}, 53(3):263--296,
  2008.

\bibitem{cadena2016past}
Cesar Cadena, Luca Carlone, Henry Carrillo, Yasir Latif, Davide Scaramuzza,
  Jos{\'e} Neira, Ian Reid, and John~J Leonard.
\newblock Past, present, and future of simultaneous localization and mapping:
  Toward the robust-perception age.
\newblock {\em IEEE Transactions on robotics}, 32(6):1309--1332, 2016.

\bibitem{calimeri2017biomedical}
Francesco Calimeri, Aldo Marzullo, Claudio Stamile, and Giorgio Terracina.
\newblock Biomedical data augmentation using generative adversarial neural
  networks.
\newblock In {\em International conference on artificial neural networks},
  pages 626--634. Springer, 2017.

\bibitem{Matterport3D}
Angel Chang, Angela Dai, Thomas Funkhouser, Maciej Halber, Matthias Niessner,
  Manolis Savva, Shuran Song, Andy Zeng, and Yinda Zhang.
\newblock Matterport3d: Learning from rgb-d data in indoor environments.
\newblock {\em International Conference on 3D Vision (3DV)}, 2017.

\bibitem{Chaplot2020ObjectGN}
Devendra~Singh Chaplot, Dhiraj Gandhi, Abhinav Gupta, and Ruslan Salakhutdinov.
\newblock Object goal navigation using goal-oriented semantic exploration.
\newblock {\em ArXiv}, abs/2007.00643, 2020.

\bibitem{ANS}
Devendra~Singh Chaplot, Dhiraj Gandhi, Saurabh Gupta, Abhinav Gupta, and Ruslan
  Salakhutdinov.
\newblock Learning to explore using active neural slam.
\newblock {\em ArXiv}, abs/2004.05155, 2020.

\bibitem{chaplot2020semantic}
Devendra~Singh Chaplot, Helen Jiang, Saurabh Gupta, and Abhinav Gupta.
\newblock Semantic curiosity for active visual learning.
\newblock In {\em European Conference on Computer Vision}, pages 309--326.
  Springer, 2020.

\bibitem{Chen2019LearningEP}
Tao Chen, Saurabh Gupta, and Abhinav Gupta.
\newblock Learning exploration policies for navigation.
\newblock {\em ArXiv}, abs/1903.01959, 2019.

\bibitem{cheng2016learning}
Gong Cheng, Peicheng Zhou, and Junwei Han.
\newblock Learning rotation-invariant convolutional neural networks for object
  detection in vhr optical remote sensing images.
\newblock {\em IEEE Transactions on Geoscience and Remote Sensing},
  54(12):7405--7415, 2016.

\bibitem{choi2021image}
Yunho Choi and Songhwai Oh.
\newblock Image-goal navigation via keypoint-based reinforcement learning.
\newblock In {\em 2021 18th International Conference on Ubiquitous Robots
  (UR)}, pages 18--21. IEEE, 2021.

\bibitem{pmlr-v48-cohenc16}
Taco Cohen and Max Welling.
\newblock Group equivariant convolutional networks.
\newblock In Maria~Florina Balcan and Kilian~Q. Weinberger, editors, {\em
  Proceedings of The 33rd International Conference on Machine Learning},
  volume~48 of {\em Proceedings of Machine Learning Research}, pages
  2990--2999, New York, New York, USA, 20--22 Jun 2016. PMLR.

\bibitem{Fastfrontier}
Anna Dai, Sotiris Papatheodorou, Nils Funk, Dimos Tzoumanikas, and Stefan
  Leutenegger.
\newblock Fast frontier-based information-driven autonomous exploration with an
  mav.
\newblock In {\em 2020 IEEE International Conference on Robotics and Automation
  (ICRA)}, pages 9570--9576, 2020.

\bibitem{dey2020group}
Neel Dey, Antong Chen, and Soheil Ghafurian.
\newblock Group equivariant generative adversarial networks.
\newblock {\em arXiv preprint arXiv:2005.01683}, 2020.

\bibitem{ICML-2016-DielemanFK}
Sander Dieleman, Jeffrey~De Fauw, and Koray Kavukcuoglu.
\newblock {Exploiting Cyclic Symmetry in Convolutional Neural Networks}.
\newblock In {\em {Proceedings of the 33rd International Conference on Machine
  Learning}}, pages 1889--1898. {JMLR.org}, 2016.

\bibitem{du2020learning}
Heming Du, Xin Yu, and Liang Zheng.
\newblock Learning object relation graph and tentative policy for visual
  navigation.
\newblock In {\em European Conference on Computer Vision}, pages 19--34.
  Springer, 2020.

\bibitem{gan2020look}
Chuang Gan, Yiwei Zhang, Jiajun Wu, Boqing Gong, and Joshua~B Tenenbaum.
\newblock Look, listen, and act: Towards audio-visual embodied navigation.
\newblock In {\em 2020 IEEE International Conference on Robotics and Automation
  (ICRA)}, pages 9701--9707. IEEE, 2020.

\bibitem{Goodfellow-et-al-2016}
Ian Goodfellow, Yoshua Bengio, and Aaron Courville.
\newblock {\em Deep Learning}.
\newblock MIT Press, 2016.
\newblock \url{http://www.deeplearningbook.org}.

\bibitem{gupta2017cognitive}
Saurabh Gupta, James Davidson, Sergey Levine, Rahul Sukthankar, and Jitendra
  Malik.
\newblock Cognitive mapping and planning for visual navigation.
\newblock In {\em Proceedings of the IEEE Conference on Computer Vision and
  Pattern Recognition}, pages 2616--2625, 2017.

\bibitem{jayaraman2018learning}
Dinesh Jayaraman and Kristen Grauman.
\newblock Learning to look around: Intelligently exploring unseen environments
  for unknown tasks.
\newblock In {\em Proceedings of the IEEE Conference on Computer Vision and
  Pattern Recognition}, pages 1238--1247, 2018.

\bibitem{Kostrikov2021ImageAI}
Ilya Kostrikov, Denis Yarats, and Rob Fergus.
\newblock Image augmentation is all you need: Regularizing deep reinforcement
  learning from pixels.
\newblock {\em ArXiv}, abs/2004.13649, 2021.

\bibitem{laskin2020reinforcement}
Michael Laskin, Kimin Lee, Adam Stooke, Lerrel Pinto, Pieter Abbeel, and
  Aravind Srinivas.
\newblock Reinforcement learning with augmented data, 2020.

\bibitem{lindeberg2021scale}
Tony Lindeberg.
\newblock Scale-covariant and scale-invariant gaussian derivative networks.
\newblock In {\em International Conference on Scale Space and Variational
  Methods in Computer Vision}, pages 3--14. Springer, 2021.

\bibitem{t2fgan}
Shuang Liu, Mete Ozay, Hongli Xu, Yang Lin, and Takayuki Okatani.
\newblock A generative model of underwater images for active landmark detection
  and docking.
\newblock In {\em 2019 IEEE/RSJ International Conference on Intelligent Robots
  and Systems (IROS)}, pages 8034--8039, 2019.

\bibitem{lv2020improving}
Yunlian Lv, Ning Xie, Yimin Shi, Zijiao Wang, and Heng~Tao Shen.
\newblock Improving target-driven visual navigation with attention on 3d
  spatial relationships.
\newblock {\em arXiv preprint arXiv:2005.02153}, 2020.

\bibitem{madani2018chest}
Ali Madani, Mehdi Moradi, Alexandros Karargyris, and Tanveer Syeda-Mahmood.
\newblock Chest x-ray generation and data augmentation for cardiovascular
  abnormality classification.
\newblock In {\em Medical Imaging 2018: Image Processing}, volume 10574, page
  105741M. International Society for Optics and Photonics, 2018.

\bibitem{mezghani2021memory}
Lina Mezghani, Sainbayar Sukhbaatar, Thibaut Lavril, Oleksandr Maksymets, Dhruv
  Batra, Piotr Bojanowski, and Karteek Alahari.
\newblock Memory-augmented reinforcement learning for image-goal navigation.
\newblock {\em arXiv preprint arXiv:2101.05181}, 2021.

\bibitem{mezghani2020learning}
Lina Mezghani, Sainbayar Sukhbaatar, Arthur Szlam, Armand Joulin, and Piotr
  Bojanowski.
\newblock Learning to visually navigate in photorealistic environments without
  any supervision.
\newblock {\em arXiv preprint arXiv:2004.04954}, 2020.

\bibitem{mnih2016asynchronous}
Volodymyr Mnih, Adria~Puigdomenech Badia, Mehdi Mirza, Alex Graves, Timothy
  Lillicrap, Tim Harley, David Silver, and Koray Kavukcuoglu.
\newblock Asynchronous methods for deep reinforcement learning.
\newblock In {\em International conference on machine learning}, pages
  1928--1937. PMLR, 2016.

\bibitem{muller2021rotation}
Philip M{\"u}ller, Vladimir Golkov, Valentina Tomassini, and Daniel Cremers.
\newblock Rotation-equivariant deep learning for diffusion mri.
\newblock {\em arXiv preprint arXiv:2102.06942}, 2021.

\bibitem{nagarajan2020learning}
Tushar Nagarajan and Kristen Grauman.
\newblock Learning affordance landscapes for interaction exploration in 3d
  environments.
\newblock {\em arXiv preprint arXiv:2008.09241}, 2020.

\bibitem{pathak2017curiosity}
Deepak Pathak, Pulkit Agrawal, Alexei~A Efros, and Trevor Darrell.
\newblock Curiosity-driven exploration by self-supervised prediction.
\newblock In {\em International conference on machine learning}, pages
  2778--2787. PMLR, 2017.

\bibitem{qi2020learning}
William Qi, Ravi~Teja Mullapudi, Saurabh Gupta, and Deva Ramanan.
\newblock Learning to move with affordance maps.
\newblock {\em arXiv preprint arXiv:2001.02364}, 2020.

\bibitem{raileanu2021automatic}
Roberta Raileanu, Max Goldstein, Denis Yarats, Ilya Kostrikov, and Rob Fergus.
\newblock Automatic data augmentation for generalization in deep reinforcement
  learning, 2021.

\bibitem{ramakrishnan2020occupancy}
Santhosh~K Ramakrishnan, Ziad Al-Halah, and Kristen Grauman.
\newblock Occupancy anticipation for efficient exploration and navigation.
\newblock In {\em European Conference on Computer Vision}, pages 400--418.
  Springer, 2020.

\bibitem{ramakrishnan2021exploration}
Santhosh~K Ramakrishnan, Dinesh Jayaraman, and Kristen Grauman.
\newblock An exploration of embodied visual exploration.
\newblock {\em International Journal of Computer Vision}, 129(5):1616--1649,
  2021.

\bibitem{Savinov2018SemiparametricTM}
Nikolay Savinov, Alexey Dosovitskiy, and Vladlen Koltun.
\newblock Semi-parametric topological memory for navigation.
\newblock {\em ArXiv}, abs/1803.00653, 2018.

\bibitem{habitat19iccv}
Manolis Savva, Abhishek Kadian, Oleksandr Maksymets, Yili Zhao, Erik Wijmans,
  Bhavana Jain, Julian Straub, Jia Liu, Vladlen Koltun, Jitendra Malik, Devi
  Parikh, and Dhruv Batra.
\newblock Habitat: {A} {P}latform for {E}mbodied {AI} {R}esearch.
\newblock In {\em Proceedings of the IEEE/CVF International Conference on
  Computer Vision (ICCV)}, 2019.

\bibitem{seifi2019look}
Soroush Seifi and Tinne Tuytelaars.
\newblock Where to look next: Unsupervised active visual exploration on 360°
  input.
\newblock {\em arXiv e-prints}, pages arXiv--1909, 2019.

\bibitem{NeuralTopological}
Devendra Singh~Chaplot, Ruslan Salakhutdinov, Abhinav Gupta, and Saurabh Gupta.
\newblock Neural topological slam for visual navigation.
\newblock In {\em 2020 IEEE/CVF Conference on Computer Vision and Pattern
  Recognition (CVPR)}, pages 12872--12881, 2020.

\bibitem{sosnovik2021scale}
Ivan Sosnovik, Artem Moskalev, and Arnold~WM Smeulders.
\newblock Scale equivariance improves siamese tracking.
\newblock In {\em Proceedings of the IEEE/CVF Winter Conference on Applications
  of Computer Vision}, pages 2765--2774, 2021.

\bibitem{sosnovik2019scale}
Ivan Sosnovik, Micha{\l} Szmaja, and Arnold Smeulders.
\newblock Scale-equivariant steerable networks.
\newblock {\em arXiv preprint arXiv:1910.11093}, 2019.

\bibitem{szot2021habitat}
Andrew Szot, Alex Clegg, Eric Undersander, Erik Wijmans, Yili Zhao, John
  Turner, Noah Maestre, Mustafa Mukadam, Devendra Chaplot, Oleksandr Maksymets,
  Aaron Gokaslan, Vladimir Vondrus, Sameer Dharur, Franziska Meier, Wojciech
  Galuba, Angel Chang, Zsolt Kira, Vladlen Koltun, Jitendra Malik, Manolis
  Savva, and Dhruv Batra.
\newblock Habitat 2.0: Training home assistants to rearrange their habitat.
\newblock 2021.

\bibitem{thiede2020the}
Erik~Henning {Thiede}, Truong~Son {Hy}, and Risi {Kondor}.
\newblock The general theory of permutation equivarant neural networks and
  higher order graph variational encoders.
\newblock {\em arXiv preprint arXiv:2004.03990}, 2020.

\bibitem{beyondfrontier}
Arnoud Visser, Xingrui-Ji, Merlijn van Ittersum, Luis~A. Gonz{\'a}lez~Jaime,
  and Lauren{\c{T}}iu~A. Stancu.
\newblock Beyond frontier exploration.
\newblock In Ubbo Visser, Fernando Ribeiro, Takeshi Ohashi, and Frank Dellaert,
  editors, {\em RoboCup 2007: Robot Soccer World Cup XI}, pages 113--123,
  Berlin, Heidelberg, 2008. Springer Berlin Heidelberg.

\bibitem{walters2020trajectory}
Robin Walters, Jinxi Li, and Rose Yu.
\newblock Trajectory prediction using equivariant continuous convolution.
\newblock {\em arXiv preprint arXiv:2010.11344}, 2020.

\bibitem{worrall2019deep}
Daniel~E. {Worrall} and Max {Welling}.
\newblock Deep scale-spaces: Equivariance over scale.
\newblock In {\em Advances in Neural Information Processing Systems},
  volume~32, pages 7364--7376, 2019.

\bibitem{BayesianMemory}
Yi Wu, Yuxin Wu, Aviv Tamar, Stuart Russell, Georgia Gkioxari, and Yuandong
  Tian.
\newblock Bayesian relational memory for semantic visual navigation.
\newblock In {\em 2019 IEEE/CVF International Conference on Computer Vision
  (ICCV)}, pages 2769--2779, 2019.

\bibitem{xiazamirhe2018gibsonenv}
Fei Xia, Amir R.~Zamir, Zhi-Yang He, Alexander Sax, Jitendra Malik, and Silvio
  Savarese.
\newblock Gibson env: real-world perception for embodied agents.
\newblock In {\em Computer Vision and Pattern Recognition (CVPR), 2018 IEEE
  Conference on}. IEEE, 2018.

\bibitem{frontierexploration}
B. Yamauchi.
\newblock A frontier-based approach for autonomous exploration.
\newblock In {\em Proceedings 1997 IEEE International Symposium on
  Computational Intelligence in Robotics and Automation CIRA'97. 'Towards New
  Computational Principles for Robotics and Automation'}, pages 146--151, 1997.

\bibitem{ye2020auxiliary}
Joel Ye, Dhruv Batra, Erik Wijmans, and Abhishek Das.
\newblock Auxiliary tasks speed up learning pointgoal navigation.
\newblock {\em arXiv preprint arXiv:2007.04561}, 2020.

\bibitem{zhang2019making}
Richard Zhang.
\newblock Making convolutional networks shift-invariant again.
\newblock In {\em International conference on machine learning}, pages
  7324--7334. PMLR, 2019.

\bibitem{target-driven}
Yuke Zhu, Roozbeh Mottaghi, Eric Kolve, Joseph~J. Lim, Abhinav Gupta, Li
  Fei-Fei, and Ali Farhadi.
\newblock Target-driven visual navigation in indoor scenes using deep
  reinforcement learning.
\newblock In {\em 2017 IEEE International Conference on Robotics and Automation
  (ICRA)}, pages 3357--3364, 2017.

\end{thebibliography}
}

\appendix

\section{More Experimental Results}

\subsection{Experimental Results of Training on MP3D}

\begin{figure}[th]
\centering
\includegraphics[width=0.7\linewidth]{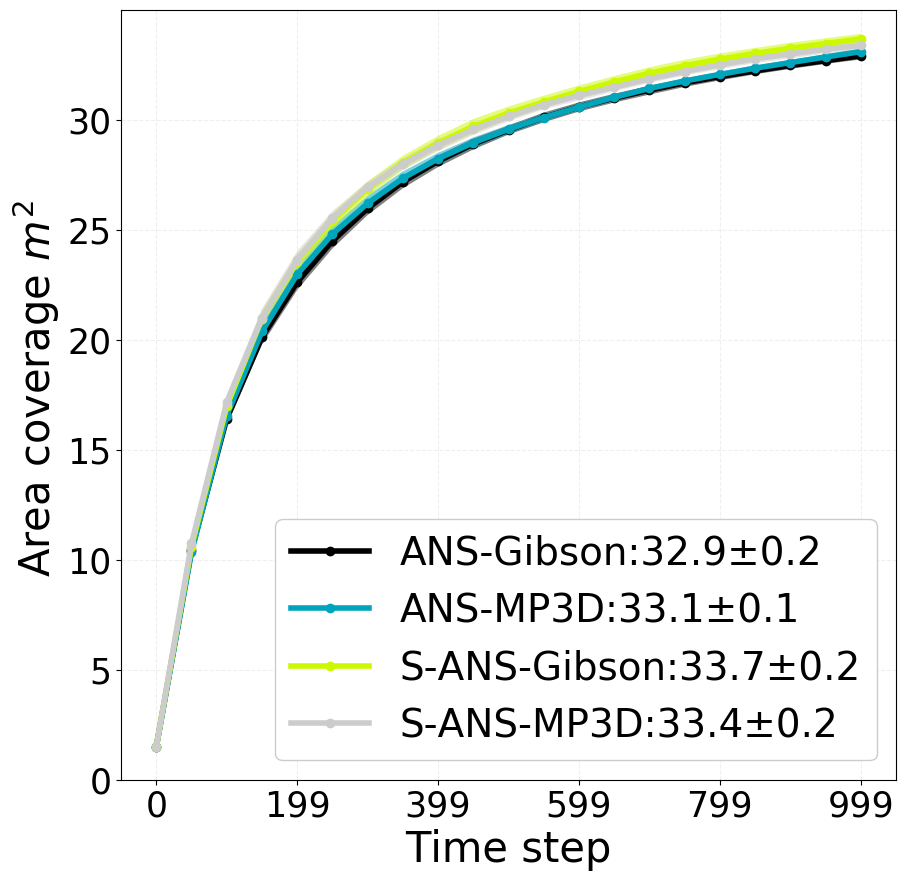}\\
\raisebox{2mm}{\small (a) Tested on Gibson}\\

\includegraphics[width=0.7\linewidth]{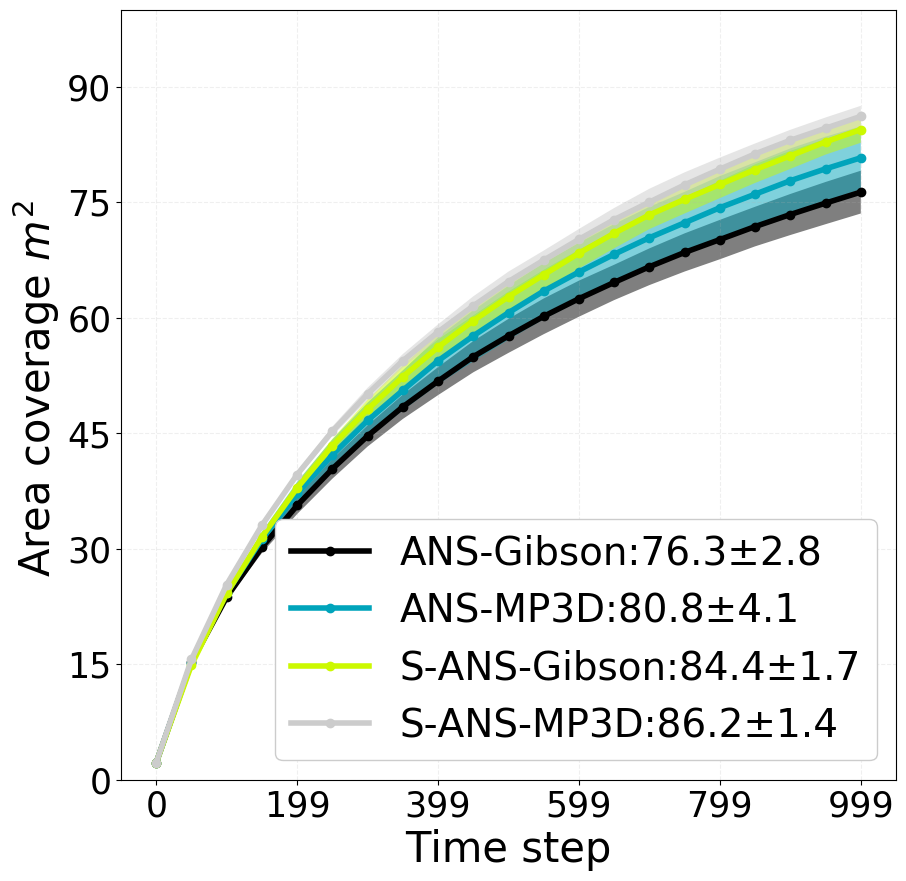}\\  
\raisebox{2mm}{\small (b) Tested on MP3D}
\vspace*{-3mm}
  \caption{Exploration performance (in area coverage, $m^2$) of ANS trained on Gibson and MP3D and S-ANS trained on Gibson and MP3D when tested on (a) Gibson and (b) MP3D. The method $\mathbb{X}$ trained on Gibson and MP3D is denoted by $\mathbb{X}$-Gibson and $\mathbb{X}$-MP3D, respectively.}
\label{fig:trainmp3d}
\end{figure}

In the main paper, we show the results of experiments in which we train the models on the Gibson dataset; the models are then tested on Gibson and the MP3D dataset. We show here the results when we train the models on MP3D and test them on Gibson and MP3D. Specifically, we evaluate two models, the original ANS and the proposed S-ANS. \cref{fig:trainmp3d} shows the results including those trained on Gibson (already shown in the main paper).  Method $\mathbb{X}$ trained on Gibson and MP3D is denoted by $\mathbb{X}$-Gibson and $\mathbb{X}$-MP3D, respectively. 

We can make the following observations. First, it is seen from Fig.~\ref{fig:trainmp3d}(b) that 
when trained and tested on MP3D, S-ANS  outperforms ANS by a large margin of $5.4m^2$ (i.e., S-ANS-MP3D = $86.2m^2$ vs. ANS-MP3D = $80.8m^2$)). This validates the effectiveness of the proposed method (i.e., S-ANS). Second, it is also seen from Fig.~\ref{fig:trainmp3d}(b) that 
the performance gap between models trained on Gibson and MP3D is smaller for S-ANS (i.e., S-ANS-Gibson = $84.4m^2$ vs. S-ANS-MP3D = $86.2m^2$) than for ANS (i.e., ANS-Gibson = $76.3m^2$ vs. ANS-MP3D = $80.8m^2$). Generally, we may consider the performance of models trained and tested on the same dataset as the upper bound of their performance. S-ANS is closer to it, supporting our conclusion that the proposed approach better handles the domain gap of the two datasets by equipping the network with the symmetries necessary for the task. 

Third, when tested on Gibson, the gaps between the models and between training datasets are small, as shown in Fig.~\ref{fig:trainmp3d}(a). Thus, the above two tendencies are not observed. We believe this is because Gibson contains smaller scenes and is simpler in complexity than MP3D. Thus, models trained on MP3D tend to achieve good performance on Gibson, e.g., ANS-Gibson = $32.9m^2$ vs. ANS-MP3D = $33.1m^2$. 


\subsection{Qualitative \& Quantitative Analysis for Invariant Representation}

We experimentally evaluate rotation invariance of the critic of S-ANS.
Specifically, we compute the standard deviation of its output and the similarity of its feature representations over inputs with different orientations. 

To compute the standard deviation of the critic's output over input rotation, we firstly sampled $Q$ state inputs $s_i, i = 1,2, \ldots Q$ of the global policy from the evaluation episodes of Gibson ($Q=1988$) and MP3D ($Q=3960$), respectively. Then, we compute a rotated state inputs set $S^*=\{s_i^k|s_i^k=r^k \cdot s_i, i\in\{1,2, \ldots, Q\},k \in \{0,1, \ldots, K-1\}\}$ for all the samples, where $r^k$ represents rotating $s_i$ by $2\pi k/K$[rad] about its center. 
Then, the standard deviation is given by
\begin{equation}
std=\frac{1}{Q}\sum_{i=1}^{Q}\sqrt{\sum_{k=0}^K(y_i^k-\bar{y_i})^2},
\label{eq:std}
\end{equation}
where $\bar{y_i}=\frac{1}{K-1}\sum_{k=0}^{K-1} y_i$ and $y_i^k = q(s_i^k)$; $q(\cdot)$ represents the function approximated by the critic. A smaller $std$ indicates better rotation invariance.

\cref{tab:criticstd} shows $std$'s of the critic of ANS and that of S-ANS (both trained on Gibson) when we set $K=24$. 
It is seen that S-ANS achieves better rotation invariance than ANS for the both test datasets. 
It is worth noting that S-ANS employs $p4$ $G$-convolution, which theoretically attains only invariance to 90 degree rotations, and has fully connected layers that are not invariant to rotation; it nevertheless achieves better invariance over $K=24$ sampling of the rotation angles. 


\begin{table}
\centering
\begin{tabular}{ccc}
\toprule
      & Gibson & MP3D  \\
\midrule
ANS   & 0.115  & 0.160  \\
S-ANS & 0.078  & 0.084 \\
\bottomrule
\end{tabular}
\caption{Rotation invariance (i.e., $std$ of (1)) of the critics of ANS and S-ANS trained on Gibson seen over the evaluation episodes of Gibson and MP3D.}
\label{tab:criticstd}
\end{table}

Next, we evaluate the similarity of the internal features of ANS and S-ANS over rotated inputs. We use the feature vector before the fully-connected layers for each model. For this purpose, we compute the similarity between two rotated inputs as
\begin{equation}
sim(\xi(s^\alpha),\xi(s^\beta))=\frac{1}{Q-1}\sum_{i=0}^{Q-1}\frac{\xi(s_i^\alpha)\cdot\xi(s_i^\beta)}{\Vert \xi(s_i^\alpha)\Vert \cdot \Vert\xi(s_i^\beta) \Vert}, 
\label{eq:cossim}
\end{equation}
where $\alpha,\beta\in\{0,1,\cdots,K-1\}$; $s^\alpha = r^\alpha\cdot s$, $s\in S^*$; and $\xi(\cdot)$ represents the function approximated by the layers before fully connected layers in the critic networks.

\begin{figure}[ht]
  \centering
  \begin{subfigure}{0.38\linewidth}
    \includegraphics[width=\textwidth]{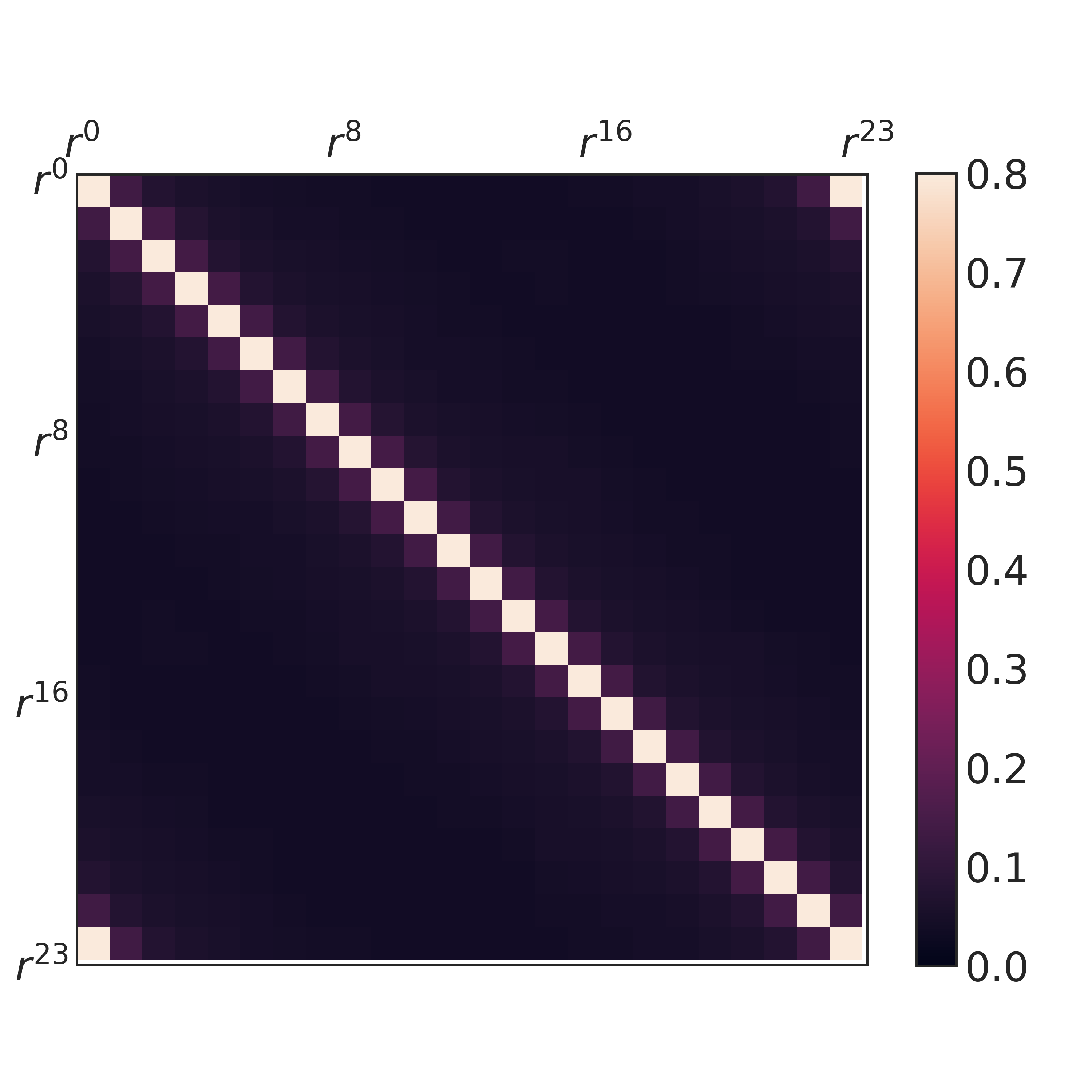}
    \vspace{-0.5cm}
    \caption{Avg.=$0.06$}
    \label{subfig:ans-gibson-cos}
  \end{subfigure}
  \begin{subfigure}{0.38\linewidth}
    \includegraphics[width=\textwidth]{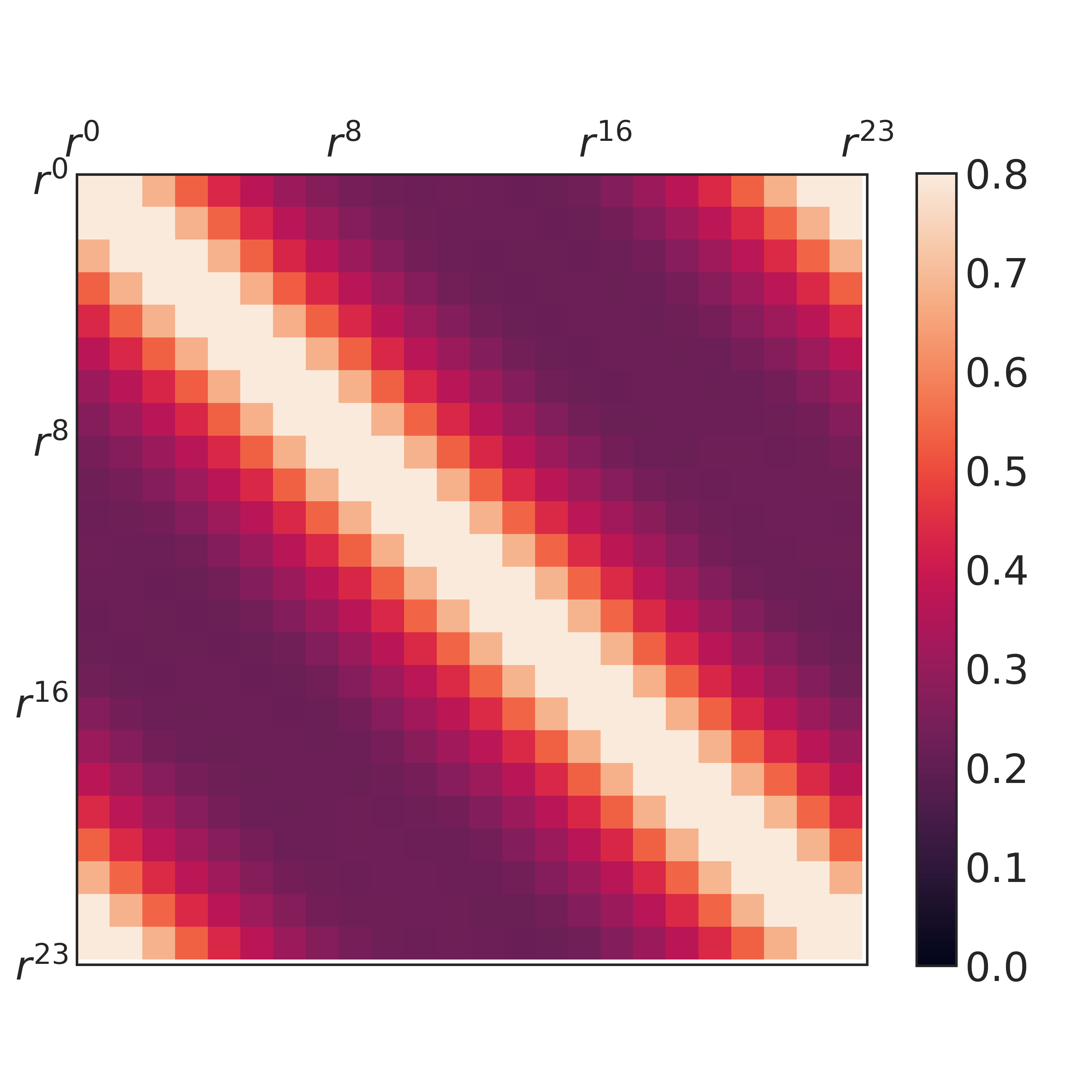}
    \vspace{-0.5cm}
    \caption{Avg.=$0.40$}
    \label{subfig:sans-gibson-cos}
  \end{subfigure}
  \begin{subfigure}{0.38\linewidth}
    \includegraphics[width=\textwidth]{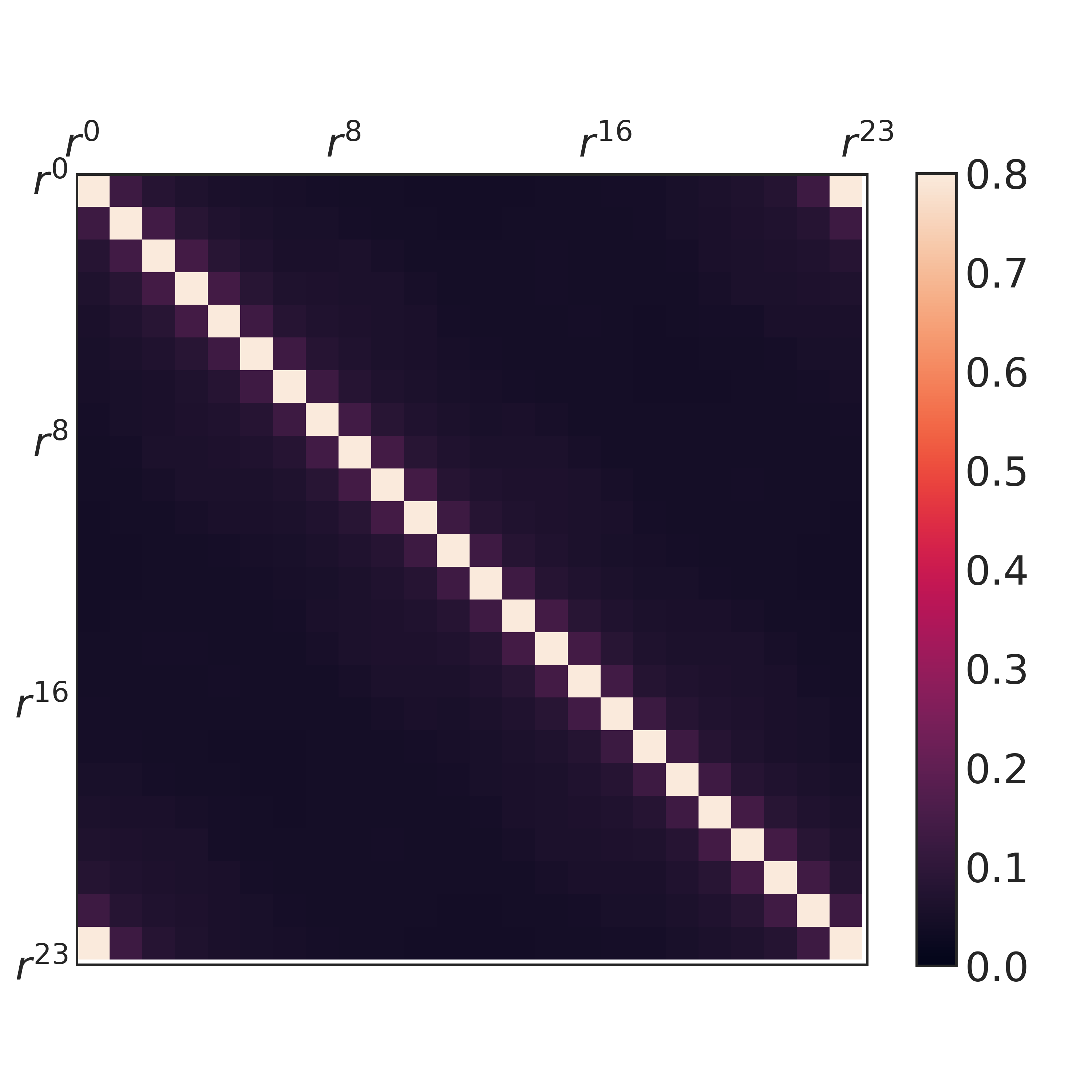}
    \vspace{-0.5cm}
    \caption{Avg.=$0.07$}
    \label{subfig:ans-mp3d-cos}
  \end{subfigure}
  \begin{subfigure}{0.38\linewidth}
    \includegraphics[width=\textwidth]{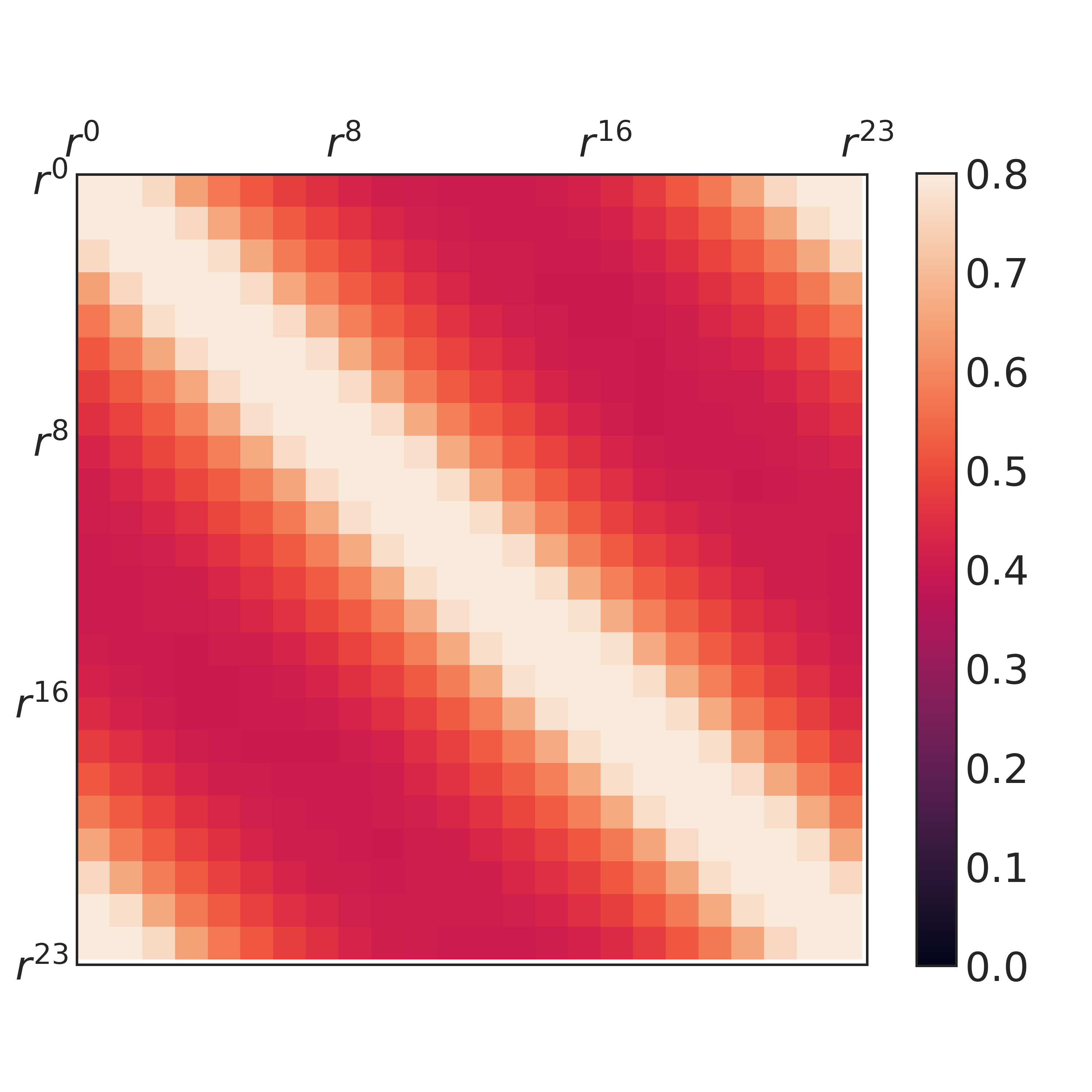}
    \vspace{-0.5cm}
    \caption{Avg.=$0.55$}
    \label{subfig:sans-mp3d-cos}
  \end{subfigure}
  \caption{Similarity of internal features over rotated inputs for (a) ANS on Gibson, (b) S-ANS on Gibson, (c) ANS on MP3D, and (d) S-ANS on MP3D. The two models are trained on Gibson. 
    Avg. indicates the mean value except the diagonal elements.}
  \label{fig:cossim}
\end{figure}

\cref{fig:cossim} shows the matrices storing the above similarity as elements for ANS and S-ANS over the evaluation episodes of Gibson and MP3D. The average similarity increases from $0.06$ of ANS to $0.40$ of S-ANS on Gibson and from $0.07$ to $0.55$ on MP3D, respectively. These verify that S-ANS achieves better rotation invariance in its feature representation.

\section{Implementation Details of FBE-RL}

This section gives implementation details of FBE-RL. FBE-RL is a RL based Frontier based exploration (FBE). It is created by combining FBE and the global policy network of ANS. Concretely, FBE-RL first computes the frontiers of the local map $h_t^l$, gaining its frontier map $m_f \in \mathbb{R}^{G\times G}$. The elements on $h_t^l$ are $0$ except for those at frontiers. Then it is combined with the map of long-term goal $m^*$, computed by the global policy network of ANS, to obtain a frontier likelihood map $m_f^{\prime} \in \mathbb{R}^{G \times G}$ by element-wise multiplication $m_f^{\prime} = m^*\odot m_f$. At last the the normalized frontier likelihood map $m_f^{\prime\prime}(x,y)$ is computed by the softmax function
\begin{equation}
    m_i^{\prime\prime}=\frac{e^{m_i}}{\sum_je^{m_j}},
\end{equation}
where $m_i$ and $m_i^{\prime\prime}$ is the $i$th element of $m_f^{\prime}$ and $m_f^{\prime\prime}$ respectively. A long-term goal is sampled from $m_f^{\prime\prime}$ for navigation.

\end{document}